\documentclass[]{article}

\usepackage{arxiv}
\usepackage{times}
\usepackage{epsfig}
\usepackage{graphicx}
\usepackage{amsmath}
\usepackage{amssymb}
\usepackage{graphicx}
\usepackage{amsmath}
\usepackage{amssymb}
\usepackage{booktabs}
\usepackage[ruled,vlined]{algorithm2e}
\usepackage{algorithmic}
\usepackage[pagebackref=true,breaklinks=true,colorlinks,bookmarks=false]{hyperref}

\usepackage{subcaption}

\usepackage[capitalize]{cleveref}
\crefname{section}{Sec.}{Secs.}
\Crefname{section}{Section}{Sections}
\Crefname{table}{Table}{Tables}
\crefname{table}{Tab.}{Tabs.}

\begin{document}

\title{EMP-SSL: Towards Self-Supervised Learning in One Training Epoch}
\author{
\centerline{
Shengbang Tong\textsuperscript{\rm 1}\thanks{Equal contribution} \quad
Yubei Chen\textsuperscript{\rm 2 *}  \quad
\textbf{Yi Ma}\textsuperscript{\rm 1,4} \quad
\textbf{Yann LeCun}\textsuperscript{\rm 2,3}
}\\
\centerline{
\textsuperscript{\rm 1}University of California, Berkeley  \quad
\textsuperscript{\rm 2}Center for Data Science, New York University\quad
}\\
\centerline{
\textsuperscript{\rm 3}Courant Inst., New York University \quad
\textsuperscript{\rm 4}Tsinghua-Berkeley Shenzhen Institute (TBSI) \quad 
}
}

\maketitle

\begin{abstract}
    Recently, self-supervised learning (SSL) has achieved tremendous success in learning image representation. Despite the empirical success, most self-supervised learning methods are rather ``inefficient'' learners, typically taking hundreds of training epochs to fully converge. In this work, we show that the key towards efficient self-supervised learning is to increase the number of crops from each image instance. Leveraging one of the state-of-the-art SSL method, we introduce a \textbf{simplistic} form of self-supervised learning method called Extreme-Multi-Patch Self-Supervised-Learning (EMP-SSL) that does not rely on many heuristic techniques for SSL such as weight sharing between the branches, feature-wise normalization, output quantization, and stop gradient, etc, and reduces the training epochs by two orders of magnitude. We show that the proposed method is able to converge to 85.1\% on CIFAR-10, 58.5\% on CIFAR-100, 38.1\% on Tiny ImageNet and 58.5\% on ImageNet-100 in just \textbf{one} epoch. Furthermore, the proposed method achieves 91.5\% on CIFAR-10, 70.1\% on CIFAR-100, 51.5\% on Tiny ImageNet and 78.9\% on ImageNet-100 with linear probing in \textbf{less than ten} training epochs. In addition, we show that EMP-SSL shows significantly better transferability to out-of-domain datasets compared to baseline SSL methods. We will release the code in https://github.com/tsb0601/EMP-SSL.

\end{abstract}
\vspace{-1em}







\section{Introduction}
\label{sec:intro}

In the past few years, tremendous progress has been made in unsupervised and self-supervised learning (SSL) \cite{lecun2022path}. Classification performance of representations learned via SSL has even caught up with supervised learning or even surpassed the latter in some cases \cite{grill2020bootstrap, chen2020simple}. This trend has opened up the possibility of large-scale data-driven unsupervised learning for vision tasks, similar to what have taken place in the field of natural language processing~\cite{brown2020language, devlin2018bert}. 

A major branch of SSL methods is joint-embedding SSL methods \cite{he2020momentum, chen2020simple, zbontar2021barlow, bardes2021vicreg}, which try to learn a representation invariant to augmentations of the the same image instance.  These methods have two goals: (1) Representation of two different augmentations of the same image should be close; (2) The representation space shall not be a collapsed trivial one\footnote{For example, all representations collapse to the same point.}, i.e., the important geometric or stochastic
structure of the data must be preserved. Many recent works \cite{chen2020simple, grill2020bootstrap, zbontar2021barlow,bardes2021vicreg} have explored various strategies and different  heuristics to attain these two properties, resulting in increasingly better performance. 

\begin{figure}[t]
  \centering
  \includegraphics[width=0.475\textwidth]{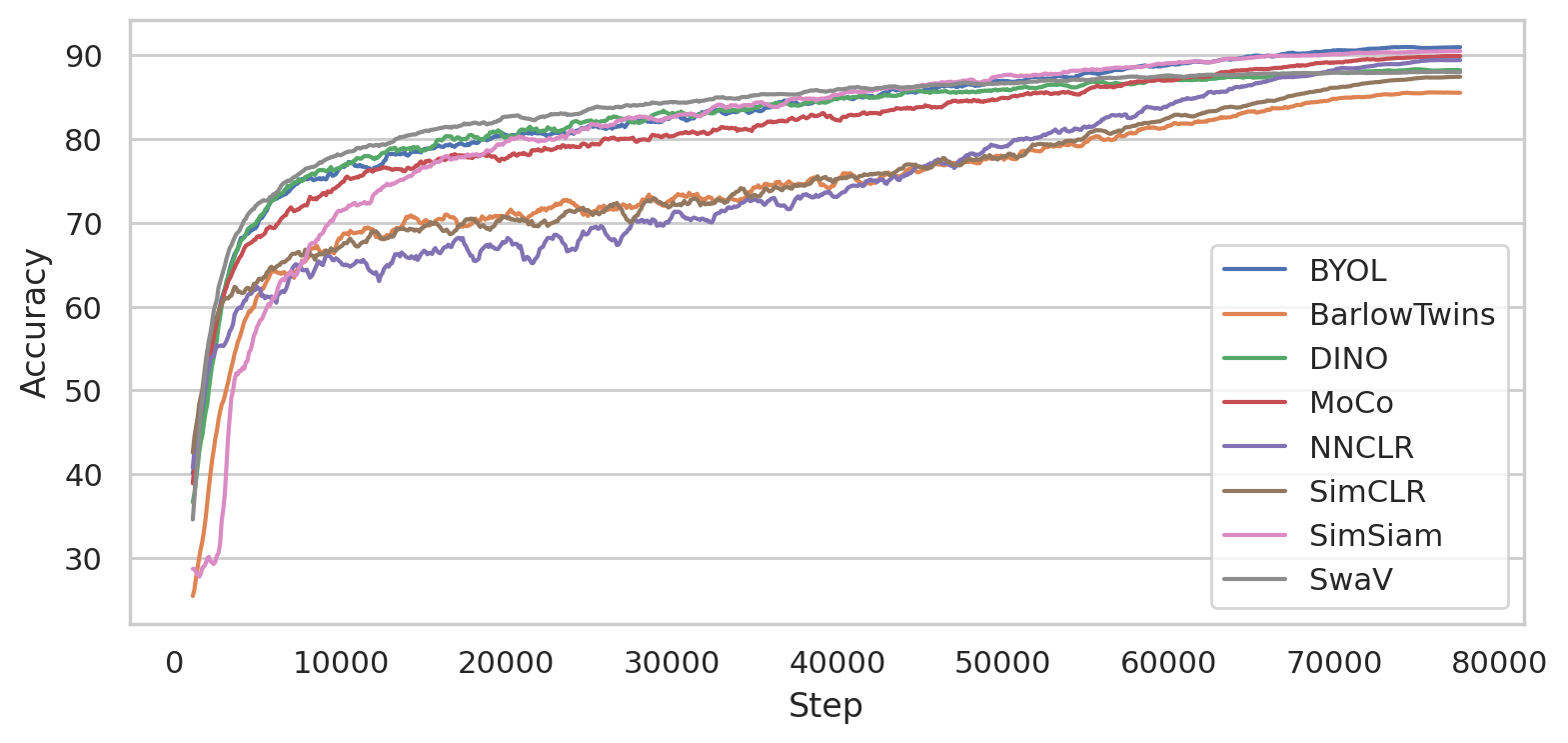}
  \caption{\textbf{The convergence plots of many SOTA SSL methods on CIFAR-10 in 800 epochs (80000 iterations).} The Accuracy of the methods is measured by k-nearest-neighbor (KNN). The plots are adopted from \cite{susmelj2020lightly}. We observe from  the plots that nearly all SOTA SSL methods take at least 500 epochs to converge to 90\%.}
  \label{fig:cifar10_benchmark}
\end{figure}

 Despite the good final performance of self-supervised learning, most of the SOTA SSL methods happen to be rather ``inefficient'' learners. Figure \ref{fig:cifar10_benchmark} plots convergence behaviors of representative SOTA SSL methods. We observe that on CIFAR-10 \cite{krizhevsky2009learning}, most methods would require at least 400 epochs to reach 90\%, whereas supervised learning typically can reach 90\% on CIFAR-10 within less than ten training epochs. The convergence efficiency gap is surprisingly large. 

While  the success of SSL has been demonstrated on a number of benchmarks, the principle or reason behind the success of this line of methods remains largely unknown. Recently, the work \cite{chen2022intra} has revealed that the success of SOTA joint-embedding SSL methods can be explained by learning distributed representation of image patches, and this discovery echos with the discovery of BagNet \cite{brendel2019approximating} in the supervised learning regime. Specifically, the work \cite{chen2022intra} show that joint-embedding SSL methods rely on successful learning the co-occurrence statistics of small image patches, and linearly aggregating of the patch representation as image representation leads to on-par or even better representation than the baseline methods. Similarly, another work based on sparse manifold transform (SMT) of small image patches \cite{chen2022minimalistic} has shown that simple white-box method can converge to close to SOTA performance in only \textit{one} epoch. Given these observations, one natural question arises: 

\textit{Can we make self-supervised learning converge faster, even in one training epoch?}

In this work, we answer this question by leveraging the observation in \cite{chen2022intra} and by pushing the number of crops in joint-embedding SSL methods to an extreme. We offer a novel new method called Extreme-Multi-Patch Self-Supervised Learning (EMP-SSL). With a simplistic formulation of joint-embedding self-supervised learning, we demonstrate that the SSL training epochs can be reduced by about  \textbf{two orders of magnitude}. In particular, we show that EMP-SSL can achieve 85.1\%  on CIFAR-10, 58.5\% on CIFAR-100, 38.1\% on Tiny ImageNet and 58.5\% on ImageNet-100 in just \textbf{one} training epoch. Moreover, with linear probing and a standard ResNet-18 backbone \cite{he2016deep}, EMP-SSL achieves 91.5\% accuracy on CIFAR-10, 70.1\% on CIFAR-100, 51.5\% on Tiny ImageNet, and 78.9\% on ImageNet-100 in less than ten training epochs. Remarkably, EMP-SSL achieves benchmark performance similar to that of SOTA methods, with more than two orders of magnitude less training epochs.




\section{The Extreme-Multi-Patch SSL Formulation} \label{sec:multi-crop}
\begin{figure*}[h]
  \centering
  \includegraphics[width=1.0\textwidth]{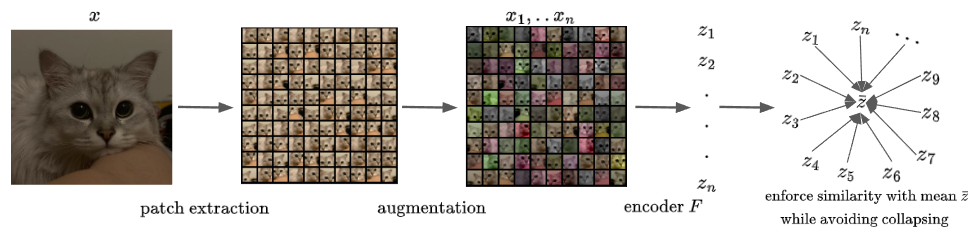}
  \vspace{-1.8em}
  \caption{\textbf{The pipeline of the proposed method.} During the training, a image is randomly cropped into $n$ fixed-size image patches with overlapping. We then apply augmentation including color jitter, greyscale, horizontal flip, gaussian blur and solarization \cite{bardes2021vicreg} to $n$ fixed-size patches. Like other SSL methods \cite{chen2020simple, bardes2021vicreg, zbontar2021barlow}, image patches are then passed into the encoder $F$ to get the representations $z$.}
  \label{fig:pipeline}
  \vspace{-1em}
\end{figure*}

\paragraph{The Overall Pipeline.} 
Like other methods for SSL \cite{chen2020simple, chen2022intra, bardes2021vicreg, zbontar2021barlow}, EMP-SSL operates on a joint embedding of augmented views of images. Inspired by the observation in \cite{chen2022intra}, the augmented views in EMP-SSL are fixed-size image patches with augmentation. As discussed in the previous sections, the purpose of joint-embedding self-supervised learning is to enforce different image patches from the same image to be close while avoiding collapsed representation. The success of these methods comes from learning patch co-occurrence \cite{chen2022intra}. In order to learn the patch co-occurrence more efficiently, we increase the number of patches in self-supervised learning to an extreme.

For a given image $x$, we break $x$ into $n$ fixed-size image patches via random crops with overlapping and apply standard augmentation identically to VICReg \cite{bardes2021vicreg} to cropped image patches get image patches $x_1, ..., x_n$. We denote $x_i$ as the $i$-th augmented image patch from $x$. For an augmented image patch $x_i$, we get embedding $h_i$ and projection $z_i$, where $h_i = f(x_i;\theta)$ and $z_i = g(h_i)$. At last, we normalize the projection $z_i$ learned. The parameter function $f(\cdot;\theta)$ is a deep neural network (ResNet-18 for example) with parameters $\theta$ and g is a much simpler neural network with only two fully connected layers. We define our encoder $F$ as $F=g(f(\cdot;\theta))$. The pipeline is illustrated as Figure \ref{fig:pipeline}. 

During the training, for a batch of $b$ images we denote as $X=[x^1, ..., x^b]$, where $x^j$ is the $j$-th image in the batch. We first augment the images as described above to get $X_1, .., X_n$ where $X_i = [x^1_i,..,x^b_i]$. Then, we pass the augmented image patches into the encoder to get the features $Z_i=F(X_i)$ and concatenate them into $Z=[Z_1,...,Z_n]$.

In this work, we adopt Total Coding Rate (TCR) \cite{ma2007segmentation, li2022neural, yu2020learning, dai2022ctrl}, which is a covariance regularization technique, to avoid collapsed representation: 
\begin{equation}
   R(Z) = \frac{1}{2}\log\det\left(I + \frac{d}{b\epsilon^{2}}{Z} {Z}^{\top}\right),
\end{equation}
where $b$ is the batch size, $\epsilon$ is a chosen size of distortion with $\epsilon>0$, and $d$ is the dimension of projection vectors. It can be seen as a soft-constrained regularization of covariance term in VICReg \cite{bardes2021vicreg}, where the covariance regularization is achieved by maximizing the Total Coding Rate (TCR). 

We would also want the representation of different image patches from the same image to be invariant, that is, different image patches from the same image should be close in the representation space. In doing so, we minimize the distance between the representation of augmented images and the mean representation of augmented images patches from the same image. Overall, the training objective is:
\begin{equation}
   \max \quad \frac{1}{n} \sum_{i=1,...,n} \Big(R(Z_i) + \lambda D(Z_i, \bar{Z})\Big),
   \label{eq:objective}
\end{equation}
where $\lambda$ is the weight for invariance loss and $\bar{Z} = \frac{1}{n}\sum_{i=1,.., n} Z_i$ is the mean of representations of  different augmented patches. In this work, we choose Cosine Similarity to implement the Distance function $D$, where $D(Z_1, Z_2)=Tr(Z_1^T Z_2)$
Hence, the larger value of $D$, the more similar $Z_i$ is to $\bar{Z}$. 
The pseudocode for EMP-SSL is shown as Algorithm \ref{algo:pseudo code}.
\definecolor{commentcolor}{RGB}{110,154,155}   
\newcommand{\PyComment}[1]{\footnotesize\ttfamily\textcolor{commentcolor}{\# #1}}  
\newcommand{\PyCode}[1]{\footnotesize\ttfamily\textcolor{black}{#1}} 

\begin{algorithm}[h]
\SetAlgoLined
    
    \PyComment{$F$: encoder network} \\
    \PyComment{lambda: weight on the invariance term} \\
    \PyComment{n: number of augmented fixed-size image patches} \\
    \PyComment{m: number of pairs to calculate invariance} \\
    \PyComment{$R$: function to calculate total coding rate} \\
    \PyComment{$D$: function to calculate cosine similarity} \\
    \PyCode{for $X$ in loader:} \\
    \Indp   
        \PyComment{augment n fixed-size image patches} \\
        \PyCode{$X_1\ldots X_n$ = \text{extract patches \& augment}($X$)} \\
        \PyCode{}\\
        \PyComment{calculate projection} \\
        \PyCode{$Z_1\ldots Z_n$ = $F(X_1)$$\ldots$$F(X_n)$} \\
        \PyCode{}\\
        \PyComment{calculate total coding rate and invariance loss} \\
        \PyCode{tcr\_loss = average([R($Z_i$) for i in range(n)]}\\
        \PyCode{inv\_loss = average([D($\bar{Z},Z_i$) for i in range(n)])}\\
        \PyCode{}\\ 
        \PyComment{calculate loss} \\
         \PyCode{loss = tcr\_loss + lambda*inv\_loss}\\
        \PyCode{}\\
        
        \PyComment{optimization step} \\
        \PyCode{loss.backward()}\\
        \PyCode{optimizer.step()}\\
        
    \Indm 
\caption{EMP-SSL PyTorch Pseudocode}
\label{algo:pseudo code}
\end{algorithm}

The objective \eqref{eq:objective} can be seen as a variant to the maximal rate reduction objective \cite{yu2020learning}, or a  generalized version of many covariance-based SSL methods such as VICReg \cite{bardes2021vicreg}, I$^2$-VICReg \cite{chen2022intra}, TCR \cite{li2022neural} and Barlow Twins \cite{zbontar2021barlow}, in which $n$ is set to 2 for the common 2-view self-supervised learning methods. In this work, we choose $n$ to be much larger in order to learn the co-occurrence between patches much faster. Details can be found in Section \ref{paragraph:cifar implementation}.

\paragraph{Bag-of-Feature Model.}  \label{paragraph:bag-of-feature}
Similar to \cite{chen2022intra, li2022neural}, we define the representation of a given image $x$ to be the average of the embedding ${h_1, ..., h_n}$ of all the image patches. It is argued by \cite{chen2022intra, appalaraju2020towards} that the representation on the embedding $h_i$ contains more equivariance and locality that lead to better performance, whereas the projection $z_i$ is more invariant. An experimental justification can be found in \cite{appalaraju2020towards, chen2022intra}, while a rigorous justification remains an open problem.

\paragraph{Architecture.} In this work, we try to adopt the simplistic form of network architecture used in self-supervised learning. 
Specifically, EMP-SSL does not require prediction networks, momentum encoders, non-differentiable operators, or stop gradients. While these methods have been shown to be effective in some self-supervised learning approaches, we leave their exploration to future work. Our focus in this work is to demonstrate the effectiveness of a simplistic yet powerful approach to self-supervised learning.


\section{Empirical Results} \label{sec:results}
In this section, we first verify the efficiency of the proposed objective in terms of convergence speed on standard datasets: CIFAR-10 \cite{krizhevsky2009learning}, CIFAR-100 \cite{krizhevsky2009learning}, Tiny ImageNet \cite{le2015tiny} and ImageNet-100 \cite{deng2009imagenet}.  We then use t-SNE maps to show that, despite only a few epochs, EMP-SSL already learns meaningful representations.  Next, we provide an ablation study on the number of patches $n$ in the objective \eqref{eq:objective} to justify the significance of patches in the convergence of our method. Finally, we present some empirical observations that the proposed method enjoys much better transferability to out-of-distribution datasets compared with other SOTA SSL methods.  

\paragraph{Experiment Settings and Datasets.}
\label{paragraph:cifar implementation} 
We provide empirical results on the standard CIFAR-10\cite{krizhevsky2009learning}, CIFAR-100 \cite{krizhevsky2009learning}, Tiny ImageNet \cite{le2015tiny} and ImageNet-100 \cite{deng2009imagenet} datasets, which contains 10, 100, 200 and 100 classes respectively. Both CIFAR-10 and CIFAR-100 contain 50000 training images and 10000 test images, size $32\times32\times3$. Tiny ImageNet contains 200 classes, 100000 training images and 10000 test images. Image size of Tiny ImageNet is $64\times 64\times 3$. ImageNet-100 is a common subset of ImageNet with 100 classes \footnote{The selection of 100 classes can be found in \cite{da2022solo}.}, containing around 126600 training images and 5000 test images, size $224\times 224$. 

For all the experiments, we use a ResNet-18 \cite{he2016deep} as the backbone and train for at most 30 epochs.  We use a batch size of 100, the LARS optimizer \cite{you2017large} with $\eta$ set to 0.005, and a weight decay of 1e-4. The learning rate is set to 0.3 and follows a cosine decay schedule with a final value 0. In the TCR loss, $\lambda$ is set to 200.0 and $\epsilon^2$ is set to 0.2. The projector network consists of 2 linear layers with respectively 4096 hidden units and 512 output units. The data augmentations used are identical to those of VICReg \cite{bardes2021vicreg}. For the number of image patches, we have set $n$ to 200 unless specified otherwise. For both CIFAR-10 and CIFAR-100, we use fixed-size image patches $16\times16$ and upsample to $32\times 32$. For Tiny ImageNet, we use a fixed patch size of $32 \times 32$ and upsample to $64\times 64$ for the convenience of using ResNet-18. For ImageNet-100, we use a fixed patch size of $112\times112$ and upsample to $224\times224$. We train an additional linear classifier to evaluate the performance of the learned representation. The additional classifier is trained with 100 epochs, optimized by SGD optimizer \cite{robbins1951stochastic} with a learning rate of 0.03. 

\paragraph{A Note on Reproducing Results of SOTA Methods.} 
We have selected five representative SOTA SSL methods \cite{chen2020simple, grill2020bootstrap, bardes2021vicreg, li2021efficient,caron2020unsupervised}
as baselines. 
For reproduction of other methods, we use sololearn \cite{da2022solo}, which is one of the best SSL libraries on github. For CIFAR-10 and CIFAR-100, we run each method 3 times for 1000 epochs with their \textit{optimal} parameters provided. For Tiny ImageNet, We notice that sololearn \cite{da2022solo} does not contain code to reproduce results on Tiny ImageNet and nearly all SOTA methods does not have official github code on Tiny ImageNet. So for fairness comparison, we adopt result from other peer-reviewed works \cite{ermolov2021whitening, zheng2021ressl}, in which SOTA methods are trained to 1000 epochs on ResNet-18. For ImageNet-100, we adopt results from sololearn \cite{da2022solo}. All baseline methods run for 400 epochs, which is commonly used for these SSL methods. 

Because our models are trained only on fixed-size image patches, we use bag-of-feature as the representation as described in Section \ref{paragraph:bag-of-feature}. Following \cite{chen2022intra}, we choose 128 as the number of patches in the bag-of-feature. The other reproduced models follow the routine in \cite{chen2020simple, he2020momentum, bardes2021vicreg} and evaluate on the whole image. We acknowledge that this may give a slight advantage to EMP-SSL. But as shown in Table 1, 2, 3 in \cite{chen2022intra}, the difference between bag-of-feature and whole image evaluation in \cite{chen2020simple, he2020momentum, bardes2021vicreg} is at most 1.5\%. We consider it negligible since this is a work about data efficiency of SSL methods, not about advancing the SOTA performance. 
    
    
    

\subsection{Self-Supervised Learning in One Epoch}
In this subsection, we conducted an experiment for one epoch and set the learning rate weight decay to one epoch, while keeping all other experiment settings the same as in \ref{paragraph:cifar implementation}. Table \ref{tab:compare_oneepoch} shows the results of our method, as well as some representative state-of-the-art (SOTA) SSL methods. From the Table, we observe that, even only seen the dataset once, the method is able to converge to a decent result close to the fully converged SOTA performance. This demonstrates great potential not only in improving the convergence of current SSL methods, but also in other fields of computer vision where the data can only be seen once, such as in online learning, incremental learning and robot learning.

\begin{table*}[h]
    \centering
    \small
    \setlength{\tabcolsep}{4.6pt}
    \renewcommand{\arraystretch}{1.25}
    \begin{tabular}{l|c|c|c|c}
    
     & \multicolumn{1}{c|}{CIFAR-10} & \multicolumn{1}{c|}{CIFAR-100} & \multicolumn{1}{c|}{Tiny ImageNet} & \multicolumn{1}{c}{ImageNet-100} \\
    
   {Methods} & 1000 Epoch & 1000 Epoch & 1000 epochs & 400 epochs  \\
    \hline
    \hline
    SimCLR                          &  0.910  &  0.662  & 0.488 & 0.776  \\
    BYOL                        &  0.926  &  0.708  & 0.510 & 0.802 \\
    VICReg          &  0.921 &  0.685  & - & 0.792   \\
    SwAV         &  0.923  &  0.658   & - &0.740 \\
    ReSSL          &  0.914   & 0.674 & - & 0.769    \\
    \hline
    EMP-SSL (\textbf{1 Epoch})  & \multicolumn{1}{c|}{0.851} & \multicolumn{1}{c|}{0.585} & 0.381 & 0.585\\
    \end{tabular}      
    \caption{\textbf{Performance of EMP-SSL with 1 epoch vs standard self-supervised SOTA methods converged.} Accuracy is measured by linear probing.} 
    \label{tab:compare_oneepoch}
\end{table*}


\subsection{Fast Convergence on Standard Datasets} \label{sec:fast_converge}
\paragraph{Comparisons with Other SSL Methods on CIFAR-10 and CIFAR-100.}

In Table \ref{tab:compare_cifar}, we present results of EMP-SSL trained up to 30 epochs and other SOTA methods trained up to 1000 epochs following the routine in \cite{chen2020simple, bardes2021vicreg, zbontar2021barlow}. On CIFAR-10, EMP-SSL is observed to converge much faster than traditional SSL methods. After just one epoch, it achieves 80.6\% accuracy with 20 patches and 82.6\% accuracy with 200 patches. In only ten epochs, it converges to more than 90\%, which is considered as the state-of-the-art result for self-supervised learning methods on CIFAR-10. By 30 epochs, EMP-SSL surpasses all current methods, achieving over 93\% accuracy as shown in the 1000 epochs column in Table \ref{tab:compare_cifar}. 

Similarly, EMP-SSL also converges very quickly on more complex datasets like CIFAR-100. In Table \ref{tab:compare_cifar}, with just 10 epochs, EMP-SSL is able to converge to 70.1\% accuracy. The method further surpasses current SOTA methods with 30 epochs of training. Due to the increased complexity of the CIFAR-100 dataset, the difference between EMP-SSL and standard SSL methods in the first 30 epochs becomes even larger than that observed on CIFAR-10.

\begin{table*}[h]
    \centering
    \small
    \setlength{\tabcolsep}{1.5pt}
    \begin{tabular}{l|cccc|cccc}
    
     & \multicolumn{4}{c|}{CIFAR-10} & \multicolumn{4}{c}{CIFAR-100}  \\
    
   {Methods} & 1 Epoch & 10 Epochs & 30 Epochs & 1000 Epochs & 1 Epoch & 10 Epochs & 30 Epochs & 1000 Epochs \\
    \hline
    \hline
    SimCLR                          &  0.282   & 0.565 & 0.663 & 0.910 &  0.054   & 0.185 & 0.341 & 0.662      \\
    BYOL                        &  0.249  & 0.489 & 0.684 & \textbf{0.926}  &  0.043  & 0.150 & 0.349 & \textbf{0.708}     \\
    VICReg          &  0.406   & 0.697 & 0.781 & 0.921  &  0.079   & 0.319 &  0.479& 0.685      \\
    SwAV         &  0.245   & 0.532 &  0.767 &  0.923    &  0.028   & 0.208 &  0.294 & 0.658      \\
    ReSSL          &  0.245   & 0.256 & 0.525& 0.914  &  0.033   & 0.122 & 0.247& 0.674       \\
    
    EMP-SSL (20 patches)  & \textbf{0.806} & \textbf{0.907} & \textbf{0.931} & -   & \textbf{0.551} & \textbf{0.678} & \textbf{0.724} & -       \\
    
    EMP-SSL (200 patches)  & \textbf{0.826} & \textbf{0.915} & \textbf{0.934} & -   & \textbf{0.577} & \textbf{0.701} & \textbf{0.733} & -       \\
    \end{tabular}      
    \caption{\textbf{Performance on CIFAR-10 and CIFAR-100 of EMP-SSL and standard self-supervised SOTA methods with different epochs.} Accuracy is measured by training linear classifier on learned embedding representation. Since EMP-SSL already converges with 10 epochs, we do not run it to 1000 epochs like other SOTA methods. Best are marked in  \textbf{bold}.}
    \label{tab:compare_cifar}
\end{table*}
%

\begin{figure}[h]
  \centering
  \begin{minipage}[b]{0.45\textwidth}
    \includegraphics[width=\textwidth]{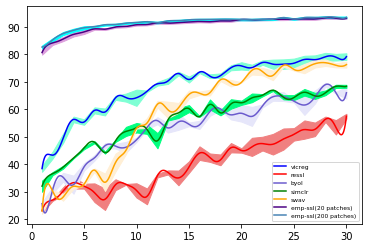}
    \caption{\textbf{The convergence plot of EMP-SSL trained on CIFAR-10 for 30 epochs.} The Accuracy is measured by linear probing. Each method runs 3 random seeds and standard deviation is displayed by shadows.}
    \label{fig:cifar10_converge}
  \end{minipage}
  \hfill
  \begin{minipage}[b]{0.45\textwidth}
    \includegraphics[width=\textwidth]{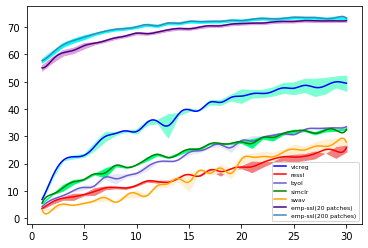}
    \caption{\textbf{The convergence plot of EMP-SSL trained on CIFAR-100 for 30 epochs.} The Accuracy is measured by linear probing. Each method runs 3 random seeds and standard deviation is displayed by shadows.}
    \label{fig:cifar100_converge}
  \end{minipage}
\end{figure}

We also present EMP-SSL's plot of convergence on CIFAR-10 in Figure \ref{fig:cifar10_converge} and on CIFAR-100 in Figure \ref{fig:cifar100_converge}. From Figures, we observe that the method indeed converges very quickly. In particular, it only takes at most 5 epochs for the method to achieve over 90\% on CIFAR-10 and over 65\% on CIFAR-100 with 200 patches and at most 8 epochs with 20 patches. More importantly, it is evident EMP-SSL converges after 15 epochs on both datatsets, around 93\% on CIFAR-10 and 72\% on CIFAR-100.

\paragraph{A Note on Time Efficiency.}
It is admittedly true that increasing number of patches in joint-embedding self-supervised learning could lead to increased training time. Here, we compare the time needed for each method to reach a prescribed performance on CIFAR. We use 90\% on CIFAR-10 and 65\% on CIFAR-100. conducting all experiments with two A100 GPUs. We present the results in Table \ref{tab:compare_time}. From the table, we observe that on CIFAR-10, EMP-SSL not only requires far fewer training epochs to converge, but also less runtime. This advantage becomes more evident on more complicated CIFAR-100 dataset. While previous methods require more epochs and, therefore, longer time to converge, EMP-SSL uses a few epochs to reach a good result. This result provides empirical evidence that the proposed method would enjoy the faster speed of training, especially with the setting with 20 patches. Beyond advantage in efficiency, one may wonder how the model learned with a few  epochs is different from previous methods learned with 1000 epochs. As we will further show in section \ref{sec:vis} and \ref{sec:transfer}, the so learned model is actually better in certain aspects.

\begin{table*}[h]
    \centering
    \small
    \renewcommand{\arraystretch}{1.00}
    \begin{tabular}{l|cc|cc}
    
     & \multicolumn{2}{c|}{CIFAR-10} & \multicolumn{2}{c}{CIFAR-100}  \\
    
   {Methods} & Time & Epochs & Time & Epochs  \\
    \hline
    \hline
    SimCLR        & 385 & 842 & 453 & 907     \\
    BYOL          & 142 & 310 & 171 & 320               \\
    VICReg       & 308 & 587  & 430 & 642    \\
    SwAV          & 162 & 150 & 264 & 241   \\
    ReSSL         & 194 & 447 & 211 & 488 \\
    
    EMP-SSL (20 patches) & \textbf{35} & 8 & \textbf{30} & 7 \\
    
    EMP-SSL (200 patches) & 142 & \textbf{5} & 112 & \textbf{4}     \\
    \end{tabular}      
    \caption{\textbf{Amount of time and epochs each method takes to reach 90\% on CIFAR-10 and 65\% on CIFAR-100.} Time is measured in minutes and best are marked in \textbf{bold}.}
    \label{tab:compare_time}
\end{table*}

%

\paragraph{Comparisons with Other SSL Methods on Tiny ImageNet and ImageNet-100}
We evaluated the performance of EMP-SSL on larger datasets, namely Tiny ImageNet and ImageNet-100. Table \ref{tab:compare_Tiny ImageNet} presents the results of EMP-SSL trained for 10 epochs on these two datasets. Even on the more challenging dataset Tiny ImageNet, EMP-SSL is still able to achieve 51.5\%, which is slightly better than SOTA methods trained with 1000 epochs. A similar result is observed on ImageNet-100. The method converges to the range SOTA performance within 10 epochs. The result shows the potential of our method in applying to data sets of larger scales.
\begin{table*}[ht]
    \centering
    \small
    \renewcommand{\arraystretch}{1.25}
    \begin{tabular}{l|cc|cc}
    
     & \multicolumn{2}{c|}{Tiny ImageNet} & \multicolumn{2}{c}{ImageNet-100}  \\
   {Methods} & Epochs & Accuracy & Epochs & Accuracy \\
    \hline
    \hline
    SimCLR                         &  1000   & 0.488 & 400 & 0.776     \\
    BYOL   &  1000  & 0.510  & 400 & \textbf{0.802}   \\
    VICReg & - & - & 400 & 0.792\\ 
    SwAV & - & - & 400 &  0.740\\
    ReSSL & - & - & 400 & 0.769 \\

    EMP-SSL (ours) & 10 & \textbf{0.515} & 10 & 0.789     \\
    \end{tabular}      
    \caption{\textbf{Performance on Tiny ImageNet and ImageNet-100 of EMP-SSL vs SOTA SSL methods at different epochs}. Best results are marked in \textbf{bold}.}
    \label{tab:compare_Tiny ImageNet}
    \vspace{-1em}
\end{table*}


\subsection{Visualizing the Learned Representation} \label{sec:vis}
\begin{figure*}[h]
  \centering
  \begin{subfigure}{0.20\linewidth}
    \includegraphics[width=1.0\textwidth]{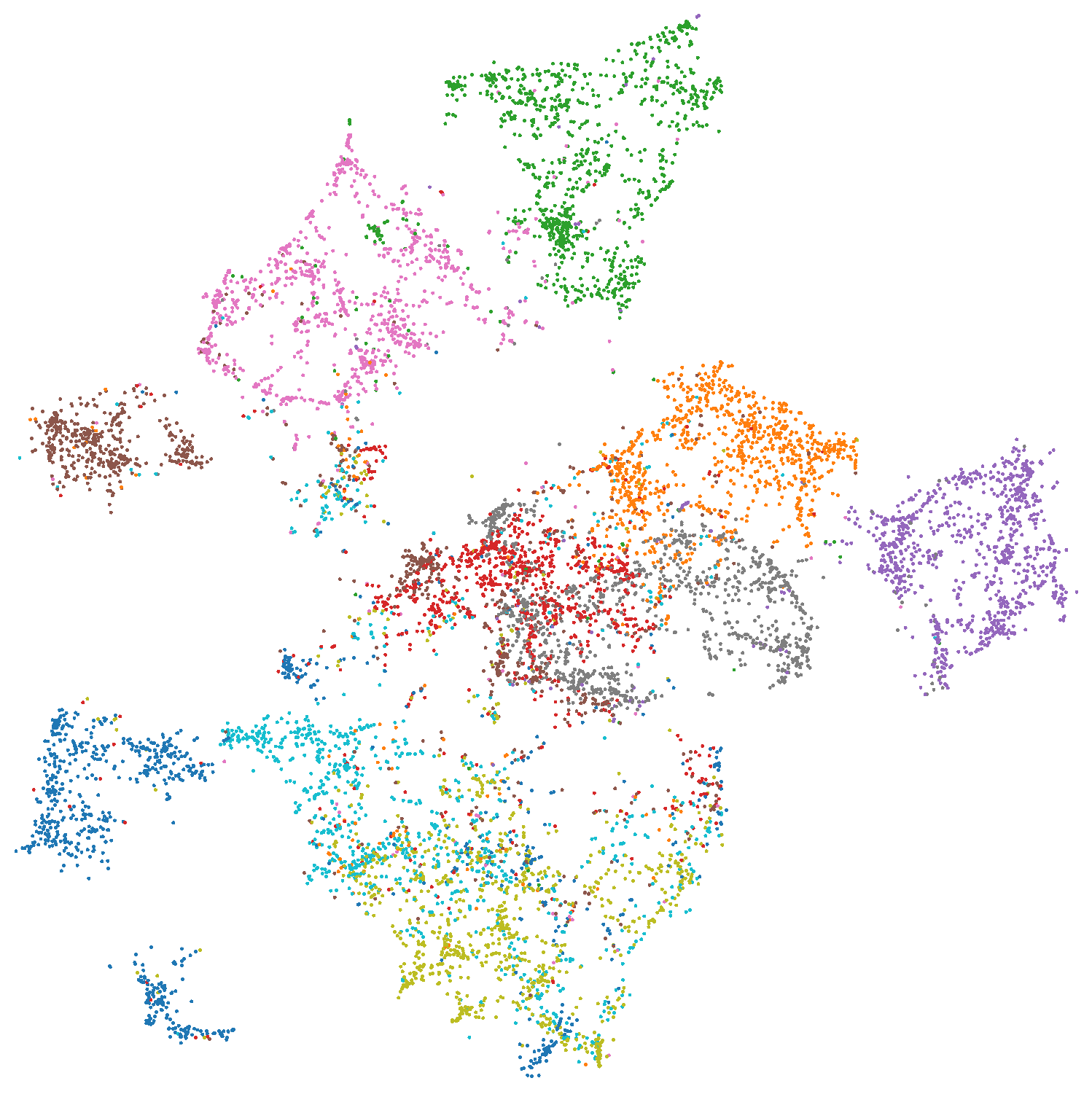}
    \caption{EMP-SSL}
    \label{fig:tsne-empvicreg}
  \end{subfigure}
  \hfill
  \begin{subfigure}{0.20\linewidth}
    \includegraphics[width=1.0\textwidth]{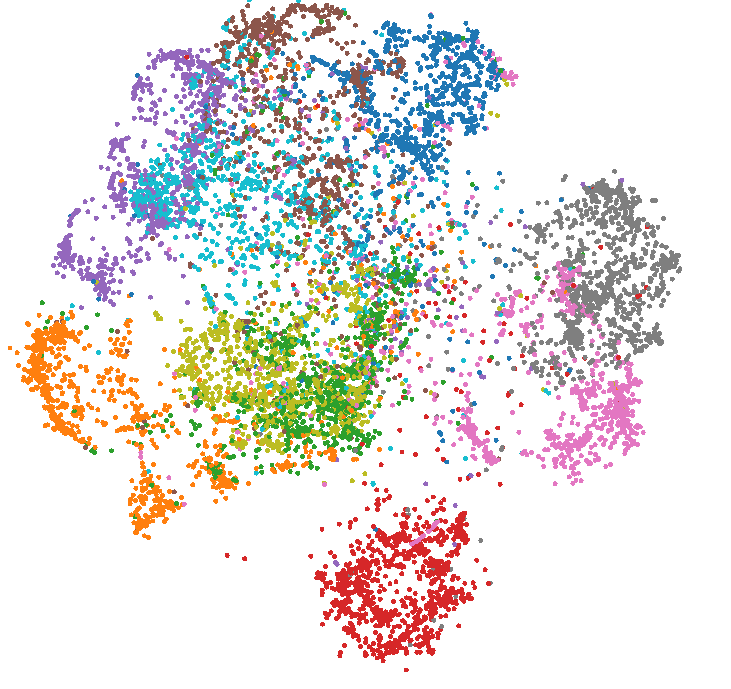}
    \caption{BYOL}
    \label{fig:tsne-byol}
  \end{subfigure}
  \hfill
  \begin{subfigure}{0.20\linewidth}
    \includegraphics[width=1.0\textwidth]{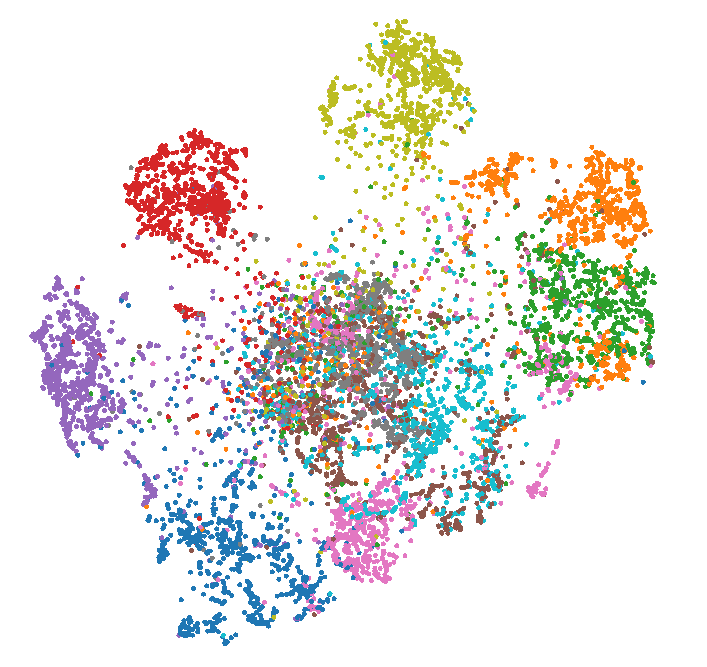}
    \caption{VICReg}
    \label{fig:tsne-vicreg}
  \end{subfigure}
  \hfill
  \begin{subfigure}{0.20\linewidth}
    \includegraphics[width=1.0\textwidth]{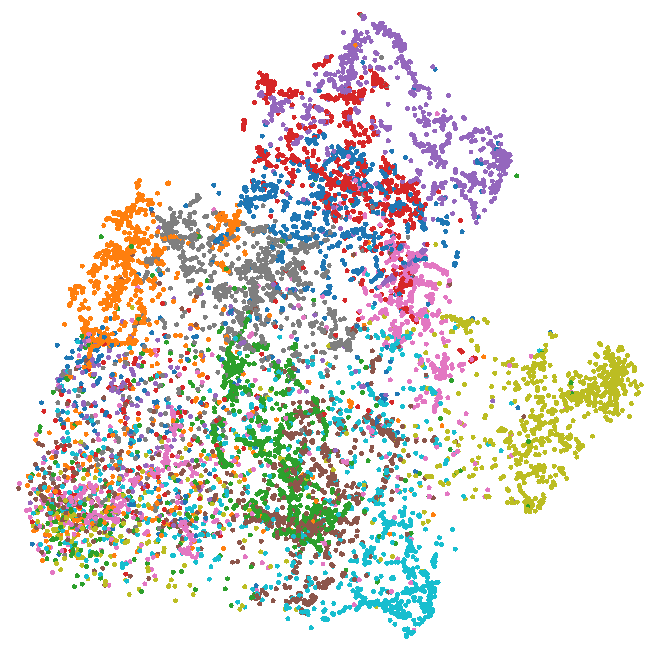}
    \caption{SwAV}
    \label{fig:tsne-swav}
  \end{subfigure}
  \caption{\textbf{t-SNE of learned representation on CIFAR-10.} We use projection vectors to generate the t-SNE graph. }
  \label{fig:cifar10_tsne}
\end{figure*}

To further understand the representations learned by EMP-SSL with a few epochs, we  visualize the features learned using t-SNE  \cite{van2008visualizing}. In Figure \ref{fig:cifar10_tsne}, we visualize the learned representations of the training set of CIFAR-10 by t-SNE. EMP-SSL is trained up to 10 epochs with 200 patches and other SOTA methods are trained up to  1000 epochs. All t-SNEs are produced with the same set of parameters. Each color represents one class in CIFAR-10. As shown in the figure, EMP-SSL learns much more separated and structured representations for different classes. Comparing to other SOTA methods, the features learned by EMP-SSL show more refined low-dim structures. For a number of classes, such as the pink, purple, and green classes, the method even learns well-structured representation inside each class. Moreover, the most amazing part is that all such structures are learned from training  with just 10 epochs!


\subsection{Ablation studies of EMP-SSL} 
\begin{figure}[h]
  \centering
  \includegraphics[width=0.4\textwidth]{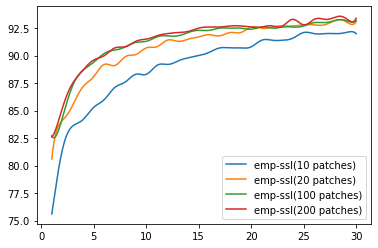}
  \caption{\textbf{Ablation Study on the number of patches $n$.} Experiments are conducted on CIFAR-10. }
  \label{fig:cifar10_ablation_n}
\end{figure}

We provide ablation studies on the number of patches $n$ to illustrate the importance of patch number in joint-embedding SSL. 
All experiments on done on CIFAR-10, with training details same with the ones in \ref{paragraph:cifar implementation}. Figure \ref{fig:cifar10_ablation_n} shows the effect that the number of patches $n$ has on the convergence and performance of EMP-SSL. As the number $n$ increases, the accuracy clearly rises sharply. Increasing number of patches $n$ used in training will facilitate the models to learn patch representation and the co-occurrence, and therefore accelerate the convergence of our model.

\subsection{Transferability to Out of Domain Data} \label{sec:transfer}

Aside from converging with much fewer epochs, we are interested in whether EMP-SSL can bring additional benefits comparing to standard 2-view self-supervised learning methods trained to 1000 epochs. In this section, we provide an interesting empirical observation: the method's better transferability to out of domain data. We conduct two sets of experiments: (1) models pretrained on CIFAR-10 and linearly evaluated on CIFAR-100 (2) models pretrained on CIFAR-100 and linearly evaluated on CIFAR-10. We present the results of these two sets of experiments in Table \ref{tab:cifar10tocifar100} and Table \ref{tab:cifar100tocifar10} respectively. In both tables, EMP-SSL is trained for 30 epochs and other self-supervised methods are trained for 1000 epochs like previous subsections. Note that despite similar names, CIFAR-10 and CIFAR-100 have very little overlap hence they are suitable for testing model's transferability. 
\begin{table*}[ht]
    \centering
    \small
    \begin{tabular}{l|c|c}
   {Methods} & CIFAR-10 & CIFAR-100 (OOD) \\
    \hline
    \hline
    SimCLR                         &  0.910   & 0.517      \\
    BYOL                         &  0.926   & 0.552     \\
    VICReg          &  0.921   & 0.515       \\
    SwAV            & 0.923    &  0.508           \\
    ReSSL          &  0.914   & 0.529       \\
    
    EMP-SSL (20 patch) & 0.931 & 0.645       \\
    EMP-SSL (200 patch) & \textbf{0.934} & \textbf{0.648}       \\
    \end{tabular}      
    \caption{\textbf{Transfer to out-of-domain data: CIFAR-10 to CIFAR-100.} We benchmark the representation of each model evaluated on CIFAR-100 by training linear classifiers on features extracted by models trained on CIFAR-10. Best results are marked in \textbf{bold}.}
    \label{tab:cifar10tocifar100}
\end{table*}

\begin{table*}[ht]
    \centering
    \small
    \begin{tabular}{l|c|c}
   {Methods} & CIFAR-100 & CIFAR-10 (OOD) \\
    \hline
    \hline
    SimCLR                         &  0.662   & 0.783      \\
    BYOL                         &  0.708   & 0.813     \\
    VICReg          &  0.685   & 0.791       \\
    SwAV          &  0.658   &   0.771     \\
    ReSSL          &  0.674   & 0.780       \\
    
    EMP-SSL (20 patch) &  \textbf{0.724} &  \textbf{0.857}       \\
    EMP-SSL (200 patch) & \textbf{0.733} & \textbf{0.859}      \\
    
    \end{tabular}      
    \caption{\textbf{Transfer to out-of-domain data: CIFAR-100 to CIFAR-10} We benchmark the representation of each model evaluated on CIFAR-10 by training linear classifiers on features extracted by models trained on CIFAR-100. Best Results are in \textbf{bold}.}
    \label{tab:cifar100tocifar10}
    \vspace{-1em}
\end{table*}
In both Table \ref{tab:cifar10tocifar100} and Table \ref{tab:cifar100tocifar10}, EMP-SSL clearly demonstrates better transferability to out of domain data. Although current state of the art methods trained with 1000 epochs have shown less transferability to out-of-domain dataset. Since the main goal of self-supervised learning is to develop data-driven machine learning on wide ranges of vision tasks, it is crucial for the self-supervised learning methods to generalize well to out-of-domain data instead of overfitting the training data. From the result shown in Table \ref{tab:cifar10tocifar100} and \ref{tab:cifar100tocifar10}, we believe this work will help advance SSL methods in such a direction.

A possible explanation for this phenomenon is that a larger number of training epochs causes the models to overfit to the training dataset. Hence, converged with only a few epochs, EMP-SSL can better avoid the curse of overfitting. We leave a more rigorous explanation for this phenomenon to future studies.

\section{More Related Works}\label{sec:related_works}
There are several intertwined quests closely related to this work. Here, we touch them briefly.

\noindent {\bf Joint-Embedding Self-Supervised Learning.}
Our work is mostly related to joint-embedding self-supervised learning. The idea of instance contrastive learning was first proposed in Wu \cite{wu2018unsupervised}
The method relies on a joint embedding architecture in which two networks are trained to produce similar embeddings for different views of the same image. The idea can trace back to Siamese network architecture which was proposed in \cite{bromley1993signature}. The main challenge to these methods is \textit{collapse} where all representations are identical, ignoring the input. To overcome this issue, there are mainly two approaches: contrastive and information maximization. 
On the branch of contrastive learning, methods search for dissimilar samples from the current branch \cite{chen2020simple} or memory bank \cite{he2020momentum}. More recently, a few methods jump out of the constraint of using contrastive samples. They exploit several tricks, such as the parameter vector of one branch being a low-pass-filtered version of the parameter vector of the other branch \cite{grill2020bootstrap}, stop-gradient operation in one of the branches \cite{chen2021exploring} and batch normalization \cite{richemond2020byol}. 

On the other line of anti-collapse methods, several simpler non-constrastive methods are proposed to avoid the collapsed representation problem. TCR \cite{li2022neural}, Barlow Twins \cite{zbontar2021barlow}, and VICReg \cite{bardes2021vicreg}  propose covariance regularization to enforce a non-collapsing solution. Our work is constructed on the basis of covariance regularization to avoid collapsed representation. 

Besides exploring ways to achieve anti-collapsing solution, SwAV \cite{caron2020unsupervised} explores \textit{multi-crop} in self-supervised learning. The work uses a mix of views with different resolutions in place of two full-resolution views. It is the first work to demonstrate that \textit{multi-view} augmentation improves the performance of SSL learning. Our work simplifies and generalizes this approach and takes it to an extreme. 

Aside from the empirical success of SSL learning, work like I$^2$-VICReg \cite{chen2022intra} digs into the principle behind these methods. The work argues that success largely comes from learning a representation of image patches based on their co-occurrence statistics in the images. In this work, we adopt this observation and demonstrate that learning the co-occurrence statistics of image patches can lead to fundamental change in the efficiency of self-supervised learning as shown in Section \ref{sec:results}. 

\vspace{0.5em}
\noindent {\bf Patch-Based Representation Learning.}
Our work is also closely related to representation learning on fixed-size patches in images. The idea of exploiting patch-level representation is first raised in the supervised setting. Bagnet \cite{brendel2019approximating} classifies an image based on the co-occurrences of small local image features without taking the spatial ordering into consideration. Note, this philosophy strongly echoes with the principle raised in \cite{chen2022intra}. The paper demonstrates that this ``bag-of-feature'' approach works very well on supervised classification tasks. Many follow-up works like SimplePatch \cite{thiry2021unreasonable} and ConvMixer \cite{trockman2022patches} have all demonstrated the power of patch representation in supervised learning. 

In unsupervised learning, some early work like Jigsaw puzzle \cite{noroozi2016unsupervised} learns patch representation via solving a patch-wise jigsaw puzzle task and implicitly uses patch representation in self-supervised learning. Gidaris \cite{gidaris2020learning} takes the ``bag-of-words'' concept from NLP and applies it into the image self-supervision task. The work raises the concept of "bag-of-patches" and demonstrates that this image discretization approach can be a very powerful self-supervision in the image domain. In the recent joint-embedding self-supervised domain, I$^2$-VICReg \cite{chen2022intra} is the first work to highlight the importance of patch representation in self-supervised learning. There's another line of self-supervised learning work \cite{bao2021beit, he2022masked} based on vision transformers, which naturally uses fixed-size patch level representation due to the structure of the vision transformers.

\vspace{0.5em}
\noindent {\bf SSL Methods Not Based on Deep Learning.}
Our work has also been inspired by the classical approaches before deep learning, especially sparse modeling and manifold learning. Some earlier works approach unsupervised learning mainly from the perspective of sparsity \cite{yu2009nonlinear, lazebnik2006beyond, perronnin2010improving}. In particular, a work focuses on lossy coding \cite{ma2007segmentation} has inspired many of the recent SSL learning methods \cite{li2022neural,chen2022intra}, as well as our work to promote covariance in the representation of data through maximizing the coding rate. Manifold learning \cite{hadsell2006dimensionality, roweis2000nonlinear} and spectral clustering \cite{schiebinger2015geometry, meilua2001random} propose to model the geometric structure of high dimensional objects in the signal space. In 2018, a work called sparse manifold transform \cite{chen2018sparse} builds upon the above two areas. The work proposes to use sparsity to handle locality in the data space to build support and construct representations that assign similar values to similar points on the support. One may note that this work already shares a similar idea with the current joint-embedding self-supervised learning in the deep-learning community. 
\section{Discussion} \label{sec:Discussion}
This paper seeks to solve the long-standing inefficient problem in self-supervised learning. We introduced EMP-SSL, which tremendously increases the learning efficiency of self-supervised learning via learning patch co-occurrence. We demonstrated that with an increased number of patches during training, the method of joint-embedding self-supervised can achieve a prescribed level of performance on various datasets, such as CIFAR-10, CIFAR-100, Tiny ImageNet, and ImageNet-100, in just one epoch.  Further, we show that the method further converges to the state-of-the-art performance in about ten epochs on these datasets. Furthermore, we show that, although converged with much fewer epochs, EMP-SSL not only learns meaningful representations but also shows advantages in tasks like transferring to out-of-domain datasets.

Our work has further verified that learning patch co-occurrence is key to the success and efficiency of SSL. This discovery opens the doors to developing even more effective and efficient self-supervised learning methods, such as uncovering the mystery behind networks used in self-supervised learning and designing more interpretable and efficient "white-box" networks for learning in an unsupervised setting. This can potentially lead to more transparent and understandable models and advance the field of self-supervised learning in various applications.

Furhter, Joint-embedding self-supervised learning has not only yielded promising results in learning more discriminative latent representations, but has also inspired the development of generative models \cite{jeong2021training, tong2022unsupervised, li2022mage}. The success of this approach has also led to significant improvements in downstream tasks such as image clustering \cite{van2020scan, li2022neural, ding2023unsupervised} and incremental learning \cite{tong2022incremental, cha2021co2l, fini2022self}. Our work builds on this foundation and has the potential to further improve downstream tasks, including online learning, with the possibility of achieving significant efficiency gains.

Lastly, adapting the proposed strategy to other methods in the field of self-supervised learning could be a promising direction for future research. While it may require careful engineering tuning to apply the strategy to other methods, the potential benefits in improving the efficiency and performance of self-supervised learning make it worth exploring. 

\clearpage

{\small

\bibliographystyle{ref}

\begin{thebibliography}{10}

\bibitem{appalaraju2020towards}
Srikar Appalaraju, Yi~Zhu, Yusheng Xie, and Istv{\'a}n Feh{\'e}rv{\'a}ri.
\newblock Towards good practices in self-supervised representation learning.
\newblock {\em arXiv preprint arXiv:2012.00868}, 2020.

\bibitem{bao2021beit}
Hangbo Bao, Li~Dong, and Furu Wei.
\newblock Beit: Bert pre-training of image transformers.
\newblock {\em arXiv preprint arXiv:2106.08254}, 2021.

\bibitem{bardes2021vicreg}
Adrien Bardes, Jean Ponce, and Yann LeCun.
\newblock {VICReg}: Variance-invariance-covariance regularization for
  self-supervised learning.
\newblock {\em arXiv preprint arXiv:2105.04906}, 2021.

\bibitem{brendel2019approximating}
Wieland Brendel and Matthias Bethge.
\newblock Approximating cnns with bag-of-local-features models works
  surprisingly well on imagenet.
\newblock {\em arXiv preprint arXiv:1904.00760}, 2019.

\bibitem{bromley1993signature}
Jane Bromley, Isabelle Guyon, Yann LeCun, Eduard S{\"a}ckinger, and Roopak
  Shah.
\newblock Signature verification using a" siamese" time delay neural network.
\newblock {\em Advances in neural information processing systems}, 6, 1993.

\bibitem{brown2020language}
Tom Brown, Benjamin Mann, Nick Ryder, Melanie Subbiah, Jared~D Kaplan, Prafulla
  Dhariwal, Arvind Neelakantan, Pranav Shyam, Girish Sastry, Amanda Askell,
  et~al.
\newblock Language models are few-shot learners.
\newblock {\em Advances in neural information processing systems},
  33:1877--1901, 2020.

\bibitem{caron2020unsupervised}
Mathilde Caron, Ishan Misra, Julien Mairal, Priya Goyal, Piotr Bojanowski, and
  Armand Joulin.
\newblock Unsupervised learning of visual features by contrasting cluster
  assignments.
\newblock {\em Advances in Neural Information Processing Systems},
  33:9912--9924, 2020.

\bibitem{cha2021co2l}
Hyuntak Cha, Jaeho Lee, and Jinwoo Shin.
\newblock Co2l: Contrastive continual learning.
\newblock In {\em Proceedings of the IEEE/CVF International conference on
  computer vision}, pages 9516--9525, 2021.

\bibitem{chen2020simple}
Ting Chen, Simon Kornblith, Mohammad Norouzi, and Geoffrey Hinton.
\newblock A simple framework for contrastive learning of visual
  representations.
\newblock In {\em International conference on machine learning}, pages
  1597--1607. PMLR, 2020.

\bibitem{chen2021exploring}
Xinlei Chen and Kaiming He.
\newblock Exploring simple siamese representation learning.
\newblock In {\em Proceedings of the IEEE/CVF Conference on Computer Vision and
  Pattern Recognition}, pages 15750--15758, 2021.

\bibitem{chen2022intra}
Yubei Chen, Adrien Bardes, Zengyi Li, and Yann LeCun.
\newblock Intra-instance {VICReg}: Bag of self-supervised image patch
  embedding.
\newblock {\em arXiv preprint arXiv:2206.08954}, 2022.

\bibitem{chen2018sparse}
Yubei Chen, Dylan Paiton, and Bruno Olshausen.
\newblock The sparse manifold transform.
\newblock {\em Advances in neural information processing systems}, 31, 2018.

\bibitem{chen2022minimalistic}
Yubei Chen, Zeyu Yun, Yi~Ma, Bruno Olshausen, and Yann LeCun.
\newblock Minimalistic unsupervised learning with the sparse manifold
  transform.
\newblock {\em arXiv preprint arXiv:2209.15261}, 2022.

\bibitem{da2022solo}
Victor Guilherme~Turrisi da~Costa, Enrico Fini, Moin Nabi, Nicu Sebe, and Elisa
  Ricci.
\newblock solo-learn: A library of self-supervised methods for visual
  representation learning.
\newblock {\em J. Mach. Learn. Res.}, 23:56--1, 2022.

\bibitem{dai2022ctrl}
Xili Dai, Shengbang Tong, Mingyang Li, Ziyang Wu, Michael Psenka, Kwan Ho~Ryan
  Chan, Pengyuan Zhai, Yaodong Yu, Xiaojun Yuan, Heung-Yeung Shum, et~al.
\newblock Ctrl: Closed-loop transcription to an ldr via minimaxing rate
  reduction.
\newblock {\em Entropy}, 24(4):456, 2022.

\bibitem{deng2009imagenet}
Jia Deng, Wei Dong, Richard Socher, Li-Jia Li, Kai Li, and Li~Fei-Fei.
\newblock Imagenet: A large-scale hierarchical image database.
\newblock In {\em 2009 IEEE conference on computer vision and pattern
  recognition}, pages 248--255. Ieee, 2009.

\bibitem{devlin2018bert}
Jacob Devlin, Ming-Wei Chang, Kenton Lee, and Kristina Toutanova.
\newblock Bert: Pre-training of deep bidirectional transformers for language
  understanding.
\newblock {\em arXiv preprint arXiv:1810.04805}, 2018.

\bibitem{ding2023unsupervised}
Tianjiao Ding, Shengbang Tong, Kwan Ho~Ryan Chan, Xili Dai, Yi~Ma, and
  Benjamin~D Haeffele.
\newblock Unsupervised manifold linearizing and clustering.
\newblock {\em arXiv preprint arXiv:2301.01805}, 2023.

\bibitem{ermolov2021whitening}
Aleksandr Ermolov, Aliaksandr Siarohin, Enver Sangineto, and Nicu Sebe.
\newblock Whitening for self-supervised representation learning.
\newblock In {\em International Conference on Machine Learning}, pages
  3015--3024. PMLR, 2021.

\bibitem{susmelj2020lightly}
Igor Susmelj Matthias Heller Philipp Wirth Jeremy Prescott Malte~Ebner et~al.
\newblock Lightly.
\newblock {\em GitHub. Note: https://github.com/lightly-ai/lightly}, 2020.

\bibitem{fini2022self}
Enrico Fini, Victor G~Turrisi Da~Costa, Xavier Alameda-Pineda, Elisa Ricci,
  Karteek Alahari, and Julien Mairal.
\newblock Self-supervised models are continual learners.
\newblock In {\em Proceedings of the IEEE/CVF Conference on Computer Vision and
  Pattern Recognition}, pages 9621--9630, 2022.

\bibitem{gidaris2020learning}
Spyros Gidaris, Andrei Bursuc, Nikos Komodakis, Patrick P{\'e}rez, and Matthieu
  Cord.
\newblock Learning representations by predicting bags of visual words.
\newblock In {\em Proceedings of the IEEE/CVF Conference on Computer Vision and
  Pattern Recognition}, pages 6928--6938, 2020.

\bibitem{grill2020bootstrap}
Jean-Bastien Grill, Florian Strub, Florent Altch{\'e}, Corentin Tallec, Pierre
  Richemond, Elena Buchatskaya, Carl Doersch, Bernardo Avila~Pires, Zhaohan
  Guo, Mohammad Gheshlaghi~Azar, et~al.
\newblock Bootstrap your own latent-a new approach to self-supervised learning.
\newblock {\em Advances in neural information processing systems},
  33:21271--21284, 2020.

\bibitem{hadsell2006dimensionality}
Raia Hadsell, Sumit Chopra, and Yann LeCun.
\newblock Dimensionality reduction by learning an invariant mapping.
\newblock In {\em 2006 IEEE Computer Society Conference on Computer Vision and
  Pattern Recognition (CVPR'06)}, volume~2, pages 1735--1742. IEEE, 2006.

\bibitem{he2022masked}
Kaiming He, Xinlei Chen, Saining Xie, Yanghao Li, Piotr Doll{\'a}r, and Ross
  Girshick.
\newblock Masked autoencoders are scalable vision learners.
\newblock In {\em Proceedings of the IEEE/CVF Conference on Computer Vision and
  Pattern Recognition}, pages 16000--16009, 2022.

\bibitem{he2020momentum}
Kaiming He, Haoqi Fan, Yuxin Wu, Saining Xie, and Ross Girshick.
\newblock Momentum contrast for unsupervised visual representation learning.
\newblock In {\em Proceedings of the IEEE/CVF conference on computer vision and
  pattern recognition}, pages 9729--9738, 2020.

\bibitem{he2016deep}
Kaiming He, Xiangyu Zhang, Shaoqing Ren, and Jian Sun.
\newblock Deep residual learning for image recognition.
\newblock In {\em Proceedings of the IEEE conference on computer vision and
  pattern recognition}, pages 770--778, 2016.

\bibitem{jeong2021training}
Jongheon Jeong and Jinwoo Shin.
\newblock Training gans with stronger augmentations via contrastive
  discriminator.
\newblock {\em arXiv preprint arXiv:2103.09742}, 2021.

\bibitem{krizhevsky2009learning}
Alex Krizhevsky, Geoffrey Hinton, et~al.
\newblock Learning multiple layers of features from tiny images.
\newblock {\em online: http://www.cs.toronto.edu/kriz/cifar.html}, 2009.

\bibitem{lazebnik2006beyond}
Svetlana Lazebnik, Cordelia Schmid, and Jean Ponce.
\newblock Beyond bags of features: Spatial pyramid matching for recognizing
  natural scene categories.
\newblock In {\em 2006 IEEE computer society conference on computer vision and
  pattern recognition (CVPR'06)}, volume~2, pages 2169--2178. IEEE, 2006.

\bibitem{le2015tiny}
Ya~Le and Xuan Yang.
\newblock Tiny imagenet visual recognition challenge.
\newblock {\em CS 231N}, 7(7):3, 2015.

\bibitem{lecun2022path}
Yann LeCun.
\newblock A path towards autonomous machine intelligence.
\newblock {\em preprint posted on openreview}, 2022.

\bibitem{li2021efficient}
Chunyuan Li, Jianwei Yang, Pengchuan Zhang, Mei Gao, Bin Xiao, Xiyang Dai,
  Lu~Yuan, and Jianfeng Gao.
\newblock Efficient self-supervised vision transformers for representation
  learning.
\newblock {\em arXiv preprint arXiv:2106.09785}, 2021.

\bibitem{li2022mage}
Tianhong Li, Huiwen Chang, Shlok~Kumar Mishra, Han Zhang, Dina Katabi, and
  Dilip Krishnan.
\newblock Mage: Masked generative encoder to unify representation learning and
  image synthesis.
\newblock {\em arXiv preprint arXiv:2211.09117}, 2022.

\bibitem{li2022neural}
Zengyi Li, Yubei Chen, Yann LeCun, and Friedrich~T Sommer.
\newblock Neural manifold clustering and embedding.
\newblock {\em arXiv preprint arXiv:2201.10000}, 2022.

\bibitem{ma2007segmentation}
Yi~Ma, Harm Derksen, Wei Hong, and John Wright.
\newblock Segmentation of multivariate mixed data via lossy data coding and
  compression.
\newblock {\em IEEE transactions on pattern analysis and machine intelligence},
  29(9):1546--1562, 2007.

\bibitem{meilua2001random}
Marina Meil{\u{a}} and Jianbo Shi.
\newblock A random walks view of spectral segmentation.
\newblock In {\em International Workshop on Artificial Intelligence and
  Statistics}, pages 203--208. PMLR, 2001.

\bibitem{noroozi2016unsupervised}
Mehdi Noroozi and Paolo Favaro.
\newblock Unsupervised learning of visual representations by solving jigsaw
  puzzles.
\newblock In {\em European conference on computer vision}, pages 69--84.
  Springer, 2016.

\bibitem{perronnin2010improving}
Florent Perronnin, Jorge S{\'a}nchez, and Thomas Mensink.
\newblock Improving the fisher kernel for large-scale image classification.
\newblock In {\em European conference on computer vision}, pages 143--156.
  Springer, 2010.

\bibitem{richemond2020byol}
Pierre~H Richemond, Jean-Bastien Grill, Florent Altch{\'e}, Corentin Tallec,
  Florian Strub, Andrew Brock, Samuel Smith, Soham De, Razvan Pascanu, Bilal
  Piot, et~al.
\newblock Byol works even without batch statistics.
\newblock {\em arXiv preprint arXiv:2010.10241}, 2020.

\bibitem{robbins1951stochastic}
Herbert Robbins and Sutton Monro.
\newblock A stochastic approximation method.
\newblock {\em The annals of mathematical statistics}, pages 400--407, 1951.

\bibitem{roweis2000nonlinear}
Sam~T Roweis and Lawrence~K Saul.
\newblock Nonlinear dimensionality reduction by locally linear embedding.
\newblock {\em science}, 290(5500):2323--2326, 2000.

\bibitem{schiebinger2015geometry}
Geoffrey Schiebinger, Martin~J Wainwright, and Bin Yu.
\newblock The geometry of kernelized spectral clustering.
\newblock {\em The Annals of Statistics}, 43(2):819--846, 2015.

\bibitem{thiry2021unreasonable}
Louis Thiry, Michael Arbel, Eugene Belilovsky, and Edouard Oyallon.
\newblock The unreasonable effectiveness of patches in deep convolutional
  kernels methods.
\newblock {\em arXiv preprint arXiv:2101.07528}, 2021.

\bibitem{tong2022unsupervised}
Shengbang Tong, Xili Dai, Yubei Chen, Mingyang Li, Zengyi Li, Brent Yi, Yann
  LeCun, and Yi~Ma.
\newblock Unsupervised learning of structured representations via closed-loop
  transcription.
\newblock {\em arXiv preprint arXiv:2210.16782}, 2022.

\bibitem{tong2022incremental}
Shengbang Tong, Xili Dai, Ziyang Wu, Mingyang Li, Brent Yi, and Yi~Ma.
\newblock Incremental learning of structured memory via closed-loop
  transcription.
\newblock {\em arXiv preprint arXiv:2202.05411}, 2022.

\bibitem{trockman2022patches}
Asher Trockman and J~Zico Kolter.
\newblock Patches are all you need?
\newblock {\em arXiv preprint arXiv:2201.09792}, 2022.

\bibitem{van2008visualizing}
Laurens Van~der Maaten and Geoffrey Hinton.
\newblock Visualizing data using t-sne.
\newblock {\em Journal of machine learning research}, 9(11), 2008.

\bibitem{van2020scan}
Wouter Van~Gansbeke, Simon Vandenhende, Stamatios Georgoulis, Marc Proesmans,
  and Luc Van~Gool.
\newblock Scan: Learning to classify images without labels.
\newblock In {\em European conference on computer vision}, pages 268--285.
  Springer, 2020.

\bibitem{wu2018unsupervised}
Zhirong Wu, Yuanjun Xiong, Stella~X Yu, and Dahua Lin.
\newblock Unsupervised feature learning via non-parametric instance
  discrimination.
\newblock In {\em Proceedings of the IEEE conference on computer vision and
  pattern recognition}, pages 3733--3742, 2018.

\bibitem{you2017large}
Yang You, Igor Gitman, and Boris Ginsburg.
\newblock Large batch training of convolutional networks.
\newblock {\em arXiv preprint arXiv:1708.03888}, 2017.

\bibitem{yu2009nonlinear}
Kai Yu, Tong Zhang, and Yihong Gong.
\newblock Nonlinear learning using local coordinate coding.
\newblock {\em Advances in neural information processing systems}, 22, 2009.

\bibitem{yu2020learning}
Yaodong Yu, Kwan Ho~Ryan Chan, Chong You, Chaobing Song, and Yi~Ma.
\newblock Learning diverse and discriminative representations via the principle
  of maximal coding rate reduction.
\newblock {\em Advances in Neural Information Processing Systems},
  33:9422--9434, 2020.

\bibitem{zbontar2021barlow}
Jure Zbontar, Li~Jing, Ishan Misra, Yann LeCun, and St{\'e}phane Deny.
\newblock Barlow twins: Self-supervised learning via redundancy reduction.
\newblock In {\em International Conference on Machine Learning}, pages
  12310--12320. PMLR, 2021.

\bibitem{zheng2021ressl}
Mingkai Zheng, Shan You, Fei Wang, Chen Qian, Changshui Zhang, Xiaogang Wang,
  and Chang Xu.
\newblock Ressl: Relational self-supervised learning with weak augmentation.
\newblock {\em Advances in Neural Information Processing Systems},
  34:2543--2555, 2021.

\end{thebibliography}
}

\clearpage

\appendix
\section{Implementation Details}
Due to the limited space in the main paragraph, we include a more detailed implementation of our method and reproduction of other methods in here. 
\subsection{Training Details of EMP-SSL}
The augmentation used follows VICReg \cite{bardes2021vicreg}. A pytorch stype pseudo code is listed below:
\begin{itemize}
    \item transforms.RandomHorizontalFlip(p=0.5)
    \item transforms.RandomApply([transforms.ColorJitter(0.4, 0.4, 0.4, 0.2)], p=0.8)
    \item transforms.RandomGrayscale(p=0.2)
    \item GBlur(p=0.1)
    \item transforms.RandomApply([Solarization()], p=0.1)
\end{itemize}
All experiments are trained with at most 4 A100 GPUs.

\subsection{Training Details of other methods}
When reproducing methods of other work, we have adopted solo-Learn \cite{da2022solo} as described in the main paragraph. We followed the optimal parameters and augmentation provided by solo-learn. A special note is that we followed the default batch size, which is 256 because it is studied in many SSL methods \cite{chen2020simple, bardes2021vicreg} that larger batch size will produce better performance.

\section{More Ablation Studies}
In this section, we present more ablation studies of EMP-SSL.
\subsection{Ablation on Batch Size}
\begin{figure}[h]
  \centering
  \includegraphics[width=0.35\textwidth]{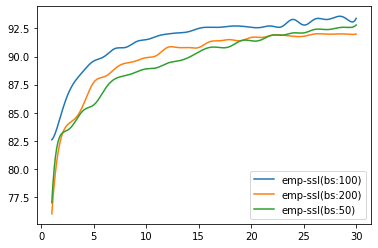}
  \caption{\textbf{Ablation Study on Batch Size} Experiments are conducted on CIFAR-10. }
  \label{fig:cifar10_bs}
\end{figure}

In this subsection, we verify if our method is applicable to different batch sizes. Again, we use CIFAR-10 to conduct ablation study and training details same in \ref{paragraph:cifar implementation}. We choose batch size of 50, 100, and 200 to conduct our ablation study. In all experiments, we use 200 patches and all the parameters are kept the same, in other words, we have not searched different hyperparameters for different batch sizes. We visualize the results of ablation study in Figure \ref{fig:cifar10_bs}. 
One may observe that batch size has little impact on the convergence of EMP-SSL. The result is very important because different batch size leads to different iteration the method has run in the same epochs. It shows that, even without changing hyperparameters, the proposed method helps the convergence of SSL method under different batch sizes.

\section{t-SNE comparison with other methods}
Due to limited space in the main text, we present the t-SNE of all of the SOTA SSL methods we have chosen to compare in here. 
\begin{figure*}[h]
  \centering
  \begin{subfigure}{0.33\linewidth}
    \includegraphics[width=1.0\textwidth]{Fig/CIFAR_10/cifar10_tsne.png}
    \caption{t-SNE of EMP-SSL}
    \label{fig:tsne_empvicreg}
  \end{subfigure}
  \hfill
  \begin{subfigure}{0.33\linewidth}
    \includegraphics[width=1.0\textwidth]{Fig/t-SNE/byol_tsne.png}
    \caption{t-SNE of BYOL}
    \label{fig:tsne_byol}
  \end{subfigure}
  \hfill
  \begin{subfigure}{0.33\linewidth}
    \includegraphics[width=1.0\textwidth]{Fig/t-SNE/vicreg_tsne.png}
    \caption{t-SNE of VICReg}
    \label{fig:tsne_vicreg}
  \end{subfigure}
  \hfill
  \begin{subfigure}{0.33\linewidth}
    \includegraphics[width=1.0\textwidth]{Fig/t-SNE/swav_tsne.png}
    \caption{t-SNE of SwAV}
    \label{fig:tsne_swav}
  \end{subfigure}
  \begin{subfigure}{0.33\linewidth}
    \includegraphics[width=1.0\textwidth]{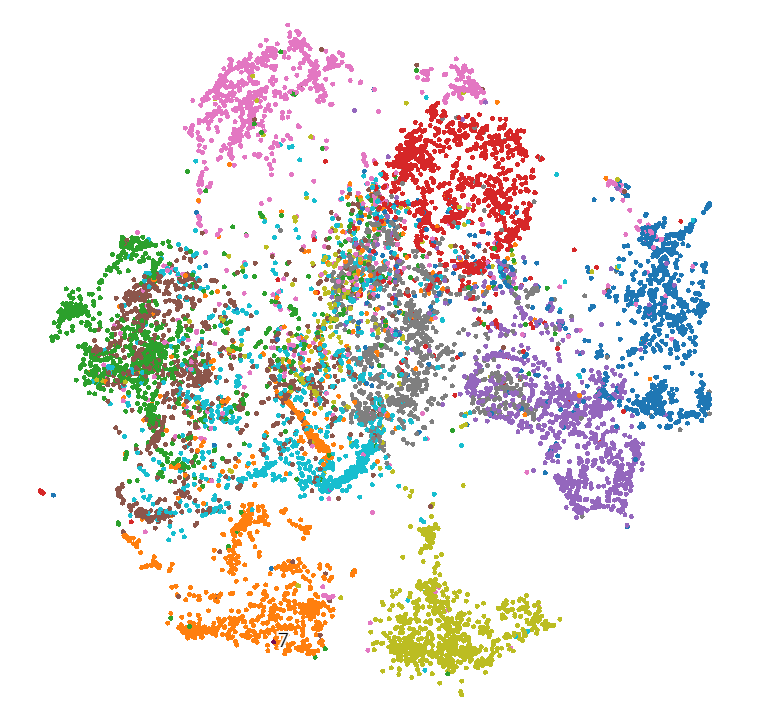}
    \caption{t-SNE of SimCLR}
    \label{fig:tsne_simclr}
  \end{subfigure}
  \begin{subfigure}{0.33\linewidth}
    \includegraphics[width=1.0\textwidth]{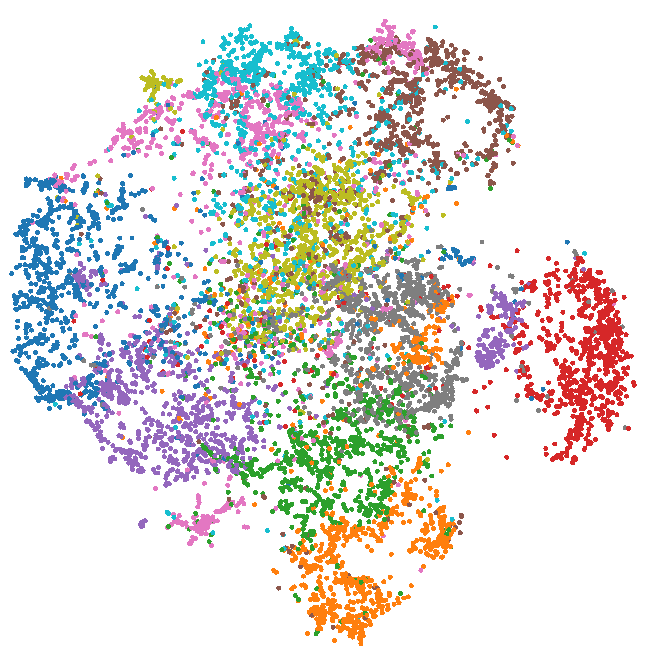}
    \caption{t-SNE of ReSSL}
    \label{fig:tsne_ressk}
  \end{subfigure}
  \caption{\textbf{t-SNE of learned representation on CIFAR-10.} We use projection vectors trained on CIFAR-10 to generate the t-SNE graph. }
  \label{fig:cifar10_tsne_more}
\end{figure*}
We present the result of all t-SNE graphs in Figure \ref{fig:cifar10_tsne_more}. Here, we draw a similar conclusion as the main paragraph, that EMP-SSL learns highly structured representation in just 10 epochs.

\end{document}